\documentclass[sigconf]{acmart}
\usepackage{amsmath,amsthm,bm,dsfont,mathtools,bbm,subcaption,hyperref,natbib}
\usepackage[normalem]{ulem} 

\usepackage{xcolor}
\usepackage{graphicx}

\usepackage{booktabs} 

\newcommand{\rmd}{\mathrm{d}}

\newtheorem{theorem1}{Theorem}[section]

\begin{document}

\title{Solving a steady-state PDE using spiking networks and neuromorphic hardware}


\setcopyright{none}
\acmConference[ICONS '20]{ICONS}{July 28--30, 2020}{Chicago, IL, USA}

\author[J.~Smith]{J.~Darby Smith}

\affiliation{%
  \department{Center for Computing Research, Sandia National Laboratories}
  \streetaddress{P.O. Box 5800, Albuquerque, NM 87185-1327}
}
\author[W.~Severa]{William Severa}

\affiliation{%
  \department{Center for Computing Research, Sandia National Laboratories}
  \streetaddress{P.O. Box 5800, Albuquerque, NM 87185-1327}
}
\author[A.~Hill]{Aaron J.~Hill}

\affiliation{%
  \department{Center for Computing Research, Sandia National Laboratories}
  \streetaddress{P.O. Box 5800, Albuquerque, NM 87185-1327}
}
\author[L.~Reeder]{Leah Reeder}

\affiliation{%
  \department{Center for Computing Research, Sandia National Laboratories}
  \streetaddress{P.O. Box 5800, Albuquerque, NM 87185-1327}
}
\author[B.~Franke]{Brian Franke}

\affiliation{%
  \department{Center for Computing Research, Sandia National Laboratories}
  \streetaddress{P.O. Box 5800, Albuquerque, NM 87185-1327}
}
\author[R.~Lehoucq ]{Richard B.~Lehoucq}

\affiliation{%
  \department{Center for Computing Research, Sandia National Laboratories}
  \streetaddress{P.O. Box 5800, Albuquerque, NM 87185-1327}
}
\author[O.~Parekh]{Ojas D.~Parekh}

\affiliation{%
  \department{Center for Computing Research, Sandia National Laboratories}
  \streetaddress{P.O. Box 5800, Albuquerque, NM 87185-1327}
}
\author[J.~Aimone]{James B.~Aimone}
\email{jbaimon@sandia.gov}
\affiliation{%
  \department{Center for Computing Research, Sandia National Laboratories}
  \streetaddress{P.O. Box 5800, Albuquerque, NM 87185-1327}
}

\begin{abstract}
The widely parallel, spiking neural networks of neuromorphic processors can enable computationally powerful formulations.  While recent interest has focused on primarily machine learning tasks, the space of appropriate applications is wide and continually expanding.  Here, we leverage the parallel and event-driven structure to solve a steady state heat equation using a random walk method.  The random walk can be executed fully within a spiking neural network using stochastic neuron behavior, and we provide results from both IBM TrueNorth and Intel Loihi implementations.  Additionally, we position this algorithm as a potential scalable benchmark for neuromorphic systems.
\end{abstract}

\maketitle

\section{Introduction}
As traditional von Neumann architectures are facing increasing scaling challenges, several beyond Moore's Law technologies offer promising and competitive advantages.  In particular, neuromorphic or neural-inspired computer architectures have seen a recent resurgence and are finding new, growing application spaces~\cite{schuman2017survey}.  While many of the proposed applications focus on size, weight and power (SWaP)-constrained computation~\cite{vineyard2019resurgence}, a high level of scalability is achievable using neural approaches, and this has lead to increasing interest in large-scale neuromorphic systems to accompany high-performance computing systems~\cite{furber2016large,monroe2014neuromorphic,hasler2013finding, aimone2018non}.  For large-scale or scientific applications, neuromorphic approaches have been applied to a number of fields and functions including cross-correlation~\cite{severa2016spiking}, dynamic programming~\cite{parekh2019dynamic}, and graph algorithms~\cite{hamilton2018neural}. 

In this paper we focus on extensions to a specific algorithm first introduced in~\cite{severa2018spiking} designed to calculate a density-based random walk process.  We leverage this existing result as a base algorithm and extend it via its application to a steady state heat equation.  Our approach has some similarities to other neural algorithms that have been used to calculate physics systems previously, for example in~\cite{lagorce2015stick}.  However, our method is differentiated by a neuromorphic-compatible random walk method applied to a steady state problem.

Additionally, we suggest that this spiking network and algorithm serves as an effective and approachable neuromorphic benchmark task.  The field is still forming best practices for benchmarking neuromorphic systems, but as new and upcoming platforms become available, it will be critical that effective benchmark tasks are developed to evaluate these systems. 

As the field develops and researches new neuromorphic applications, we expand the potential workload of neuromorphic systems from a seemingly one-trick-pony to a multi-use co-processor with well-defined advantages.  We remark that while applications are, of course, critical, theoretical understanding and characterization of these systems is equally important.  In this paper, we will provide some theoretical justification for our approach, but a full theoretical characterization of complexity is beyond the scope of this paper.  Some references on spiking network complexity are~\cite{kwisthout2020computational, verzi2018computing}.

This paper is organized as follows.  First, we provide a short introduction to the density-based random walk method and random walk methods in general in Section~\ref{rw_intro}. A mathematical description of our physics application and its probabilistic interpretation follow in Section~\ref{steady_state_section}.  Details on our neuromorphic implementation are presented in Section~\ref{neuro_implementation}.  Results from simulation and on-hardware are provided in Section~\ref{sim_results} and Section~\ref{hardware_results} respectively. In Section~\ref{eval_intro}, we discuss the current state of benchmarking for neuromorphic systems and the appropriateness of the algorithm presented herein as a benchmark for current and future neuromorphic systems.

\subsection{Random Walk Methods}
\label{rw_intro}
Probabilistic methods have a celebrated history among multiple disciplines.  In particular, random walk methods are harnessed to provide solutions in a variety of fields including computer science, physics, and operations research \cite{MASUDA20171}. A well-known example is the construction of the Black-Scholes equation in financial option pricing \cite{black1973pricing}.

Random walk methods stem from and are inspired by the basic connection between Brownian motion and the heat equation.  Consider the one-dimensional heat equation initial value problem:
\begin{align*}
\frac{\partial}{\partial t}u &= \frac{1}{2}\frac{\partial^2}{\partial x^2}u,\\
u(0,x) & = f(x).
\end{align*}
Here $f(x)$ describes the initial distribution of temperature.  Consider a Brownian motion, $W_t$.  For a fixed $t$, $W_t$ is a random variable with probability density function given by
\begin{equation*}
p(t,x,y) = \frac{1}{\sqrt{2\pi t}}\exp\left(-\frac{(y-x)^2}{2t}\right).
\end{equation*}
But note that
\begin{align*}
\mathbb{E}\left[f\left(W_t\right)\,\middle|\,W_0 = x\right] &= \int f(y)p(t,x,y)\,\rmd y\\
&= \frac{1}{\sqrt{2\pi t}}\int f(y)\exp\left(-\frac{(y-x)^2}{2t}\right)\,\rmd y,
\end{align*}
which is exactly the known solution to the heat equation.  Hence
\begin{equation*}
u(t,x) = \mathbb{E}\left[f\left(W_t\right)\,\middle|\,W_0 =x\right].
\end{equation*}

This equation allows us to directly link the solution of the heat equation to a Monte Carlo sampling method.  Specifically, paths of the process $W_t$ are sampled (i.e., random walks are sampled), the paths are evaluated at the function $f$, and then these results are averaged to obtain an approximation for the solution.  To lend a physical heuristic to this probabilistic equation, the random walks can be thought of as the paths of ``heat particles.''  Their density at a given location is related to the temperature at that location.

The density-based algorithm presented in~\cite{severa2018spiking} approaches the random walk computation by associating each node or area in the given space with a sub-circuit of neurons.  Walkers are then represented by spikes which travel from node to node or equivalently from sub-circuit to sub-circuit.  This sub-circuit has several key functional components each of which is designed using a leaky-integrate-and-fire neuron model:
\begin{itemize}
\item Counter---Represents the count of walkers at that location at a given time
\item Probability Gate---Computes the appropriate random draw deciding the direction of the walker
\item Output Gates---Sends the walkers to the connected neighbors
\item Buffer---Collects the incoming walkers as a staging area before the counter
\end{itemize}

\section{Random Walks for the Steady State Heat Equation}
\label{steady_state_section}

\subsection{Problem Statement}
Consider a thin wire of length $\ell$ meters.  At position $x=0$, the wire is attached to a cool external wash ($T=0$ degrees) such that the heat gradient is zero.  Somewhere else in the space, a heat source gives a heat-flux density $q(x)$ W$/$m\textsuperscript{3}.  The thin wire is assumed to have a heat capacity $k$ measured in W$/$m$\cdot$degrees.  Then, the steady state temperature of the wire at position $x$ is given by
\begin{align*}
-ku''(x) &=q(x),\quad x\in[0,\ell],\\
u(0)&=0,\\
u'(0)&=0.
\end{align*}
Suppose we divide through by the constant $k$, absorbing it into the quantity $q(x)$ (now measured in degrees$/$m\textsuperscript{3}).  Take $q(x)=-F\left(\ell-x\right)$ for some positive value $F$.  Physically, the interpretation is that the gradient of the heat source is $-F$ at the right endpoint of the wire and decays linearly towards the left endpoint.  We rewrite this specific problem as:
\begin{align}
\begin{split}
0&=\frac{\rmd^2}{\rmd x^2}u-F\left(\ell-x\right),\quad x\in[0,\ell],\\
u(0)&=0,\\
u'(0)&=0.
\end{split}
\label{eq:heat_problem}
\end{align}
We will focus on this specific 1D steady state heat problem. This second-order ODE can easily be solved for an analytic solution:
\begin{equation}
u(x) = \frac{F\ell x^2}{2}-\frac{Fx^3}{6}.
\label{eq:heat_ana_sol}
\end{equation}

\subsection{Probabilistic Interpretation}

We would like to recapture the analytic solution using a random walk.  To do this, we need a probabilistic interpretation of \eqref{eq:heat_problem}.  We appeal to the following theorem from~\citet{grigoriu2013}.
\begin{theorem1}{}
Let $\mathbf{x} = (x_1,\ldots,x_d)\in \mathbb{R}^d$.  Consider the PDE
\begin{align}
\begin{split}
0&=\sum_{i=1}^d \alpha_i(\mathbf{x})\frac{\partial u(\mathbf{x})}{\partial x_i} +  \frac{1}{2}\sum_{i,j=1}^d\beta_{ij}(\mathbf{x}) \frac{\partial^2u(\mathbf{x})}{\partial x_i\partial x_j} + p(\mathbf{x}),\quad\mathbf{x}\in D\subset\mathbb{R}^d\\
u(\mathbf{x})& = \xi\left(\mathbf{x}\right),\quad \mathbf{x} \in\partial D.
\end{split}
\label{eq:time_invariant_pde}
\end{align}
Let $\boldsymbol\alpha=\left(\alpha_i\right)$ and let $\mathbf{W}_t$ be a $d$-dimensional white noise process.  Suppose:
\begin{itemize}
\item $\alpha_i$, $\beta_{ij}$, and $p$ are all real-valued functions defined on $D$ and $d\in\mathbb{N}\setminus\{0\}$;
\item the matrix $\boldsymbol\beta = \left(b_{ij}\right)$ is a symmetric positive definite matrix for each $\mathbf{x}\in \mathbb{R}^d$;
\item there exists a matrix $\boldsymbol\sigma(\mathbf{x})$ such that $\boldsymbol\beta=\boldsymbol\sigma\boldsymbol\sigma^\top$;
\item there exists a process $\mathbf{X}_t$ satisfying
\begin{equation*}
\rmd \mathbf{X}_t = \boldsymbol\alpha\left(\mathbf{X}_t\right)\rmd t+\boldsymbol\sigma\left(\mathbf{X}_t\right)\rmd \mathbf{W}_t
\label{eq:sde_ti}
\end{equation*}
for each initial condition $\mathbf{X}_0 = \mathbf{x}$ with $\mathbf{x}\in D$;
\item the function $\xi$ is continuous in $\partial D$;
\item $p$ is H{\"o}lder continuous in $D$;
\item the boundaries of $D$ are regular;
\item and the partial derivatives of $u$ are bounded in $D$.
\end{itemize}
Define the random variable
\begin{equation*}
T = \inf\left\{t>0\,\middle|\,X_t\not\in D\right\}.
\label{eq:stop_time}
\end{equation*}
Then the solution to \eqref{eq:time_invariant_pde} is
\begin{equation}
u\left(\mathbf{x}\right) = \mathbb{E}\left[\xi\left(\mathbf{X}(T)\right) + \int_0^Tp\left(\mathbf{X}(s)\right)\,\rmd s\middle|\,\mathbf{X}_0 = \mathbf{x}\right].
\label{eq:ti_sol}
\end{equation}
\end{theorem1}
The assumptions in the theorem impose appropriate continuity, existence, and smoothness conditions on the functions $\boldsymbol\alpha$, $\boldsymbol\beta$, $\boldsymbol\sigma$, $p$, $\xi$, the derivatives of $u$, and their domains to ensure that both $\mathbf{X}_t$ exists for each initial condition and that $u$ exists and is unique.

This theorem does not directly apply to \eqref{eq:heat_problem} because it has mixed boundary conditions.  However, it can be modified to work in this scenario.

First note that the condition $u'(0)=0$ can be absorbed into the process $X_t$.  This condition says that the flow of heat at $x=0$ must be zero.  Heuristically, we would say that the flux of our ``heat particles'' at zero must be zero.  Therefore any sample of the random process $X_t$ must not cross zero.  Accordingly, we require $X_t$ to be a process reflecting at zero.

Next we consider the term $\mathbb{E}\left[\xi\left(\mathbf{X}(T)\right)\middle|\,\mathbf{X}_0 = \mathbf{x}\right]$ from \eqref{eq:ti_sol}.  In our scenario, the boundary condition is only given at zero and we have ensured the process $X_t$ will never exit at zero by making it a reflective process. There is no condition given for $x=\ell$, so we cannot evaluate this term.  However, we must enforce $u(0)=0$.  To do this we define
\begin{equation*}
u^*_0 = \mathbb{E}\left[-\int_0^TF\left(\ell-X(s)\right)\,\rmd s\,\middle|\,X_0 = 0\right].
\end{equation*}
Then, the probabilistic solution to \eqref{eq:heat_problem} is
\begin{equation}
u(x) = \mathbb{E}\left[-\int_0^TF\left(\ell-X(s)\right)\,\rmd s\,\middle|\,X_0 = x\right] - u^*_0,
\label{eq:heat_prob_sol}
\end{equation}
where $X_t$ is a random process reflecting at zero with law
\begin{equation*}
\rmd X_t = \sqrt{2}\,\rmd W_t
\label{eq:law_process}
\end{equation*}
elsewhere.

We take a brief moment to address the random variable $T$ for this specific problem.  $T$ is interpreted as the stopping time or absorption time of the process $X_t$.  If we knew the distribution of $T$ for the process $\left.X_t\,\middle|\,X_0=y\right.$,  then we could infer how long any simulation would need to run.  Define $\tau(y) = \mathbb{E}\left[T\,\middle|\,X_0 = y\right]$.  Let the \textit{survival probability}, or the probability that the process $\left.X_t\,\middle|\,X_0=y\right.$ has not yet left the interval $[0,\ell]$ by time $t$, be given by $S_p(y,t)$.  Then $\mathbb{P}\left[T\leq t\,\middle|\,X_0=y\right] = 1-S_p(y,t)$.  From this relation and the (backward) Fokker-Planck equation, it can be shown that
\begin{equation}
\tau(y) = \mathbb{E}\left[T\,\middle|\,X_0 = y\right] = \frac{\ell^2-y^2}{2}.
\label{eq:expected_stopping_time}
\end{equation}
While this gives us the mean of random variable $T$, the full distribution is not a straightforward calculation.

\section{Neuromorphic Implementation}
\label{neuro_implementation}

To numerically approximate via \eqref{eq:heat_prob_sol}, for each $x$ we would need to sample a large number of paths of the process $X_t$ beginning at $X(0)=x$, and average the values $-\int_0^TF\left(\ell-X(s)\right)\,\rmd s$ obtained for each path.  We cannot sample continuous paths of the process $X_t$; we must discretize in time.  We further discretize in space in order to develop a grid for a random walk.  This combination yields a Markov Chain approximation for samples of the process $X_t$.

Divide the interval of length $\ell$ into $N$ equal divisions, each of length $\Delta x$.  We construct a Markov process where random walkers move between the midpoints of these divisions according to the law of $X_t$.  For a division of time $\Delta t$, $X_{t+\Delta t}-X_t \sim \mathcal{N}\left(0,2\Delta t\right)$.  Hence we will use the Gaussian distribution to inform transition probabilities.  To simplify the random walk, we will only allow random walkers to move to an adjacent point or back to themselves.  Define
\begin{align*}
\begin{split}
p_s & = \mathbb{P}\left[-\frac{1}{2}\Delta x < X_{\Delta t} < \frac{1}{2}\Delta x\,\middle|\,X_0 =0\right]\\
&=\frac{1}{2\sqrt{\Delta t \pi}}\int_{-\frac{1}{2}\Delta x}^{\frac{1}{2}\Delta x}\exp\left(-\frac{x^2}{4\Delta t}\right)\,\rmd x,
\end{split}
\end{align*}
and
\begin{align*}
\begin{split}
p_g & = \mathbb{P}\left[X_{\Delta t}\leq-\frac{1}{2}\Delta x\,\middle|\,X_0=0\right]\\
&=\frac{1}{2\sqrt{\Delta t\pi}}\int_{-\infty}^{-\frac{1}{2}\Delta x}\exp\left(-\frac{x^2}{4\Delta t}\right)\,\rmd x.
\end{split}
\end{align*}
By the symmetry and translation invariance of the Gaussian distribution, analogous probabilities centered at different points will be equal to these two values.  Since we are restricting movement only to the nearest neighbors, it is important to choose $\Delta t$ so that you can be reasonably sure that a random walker would only move to an adjacent point or back to itself.  That is
\begin{equation*}
2\mathbb{P}\left[X_{\Delta t}\leq-\frac{3}{2}\Delta x\,\middle|\,X_0 = 0\right]<c,
\end{equation*}
for some threshold probability $c$.

\begin{figure}[h]
\centering
\includegraphics[width=3in]{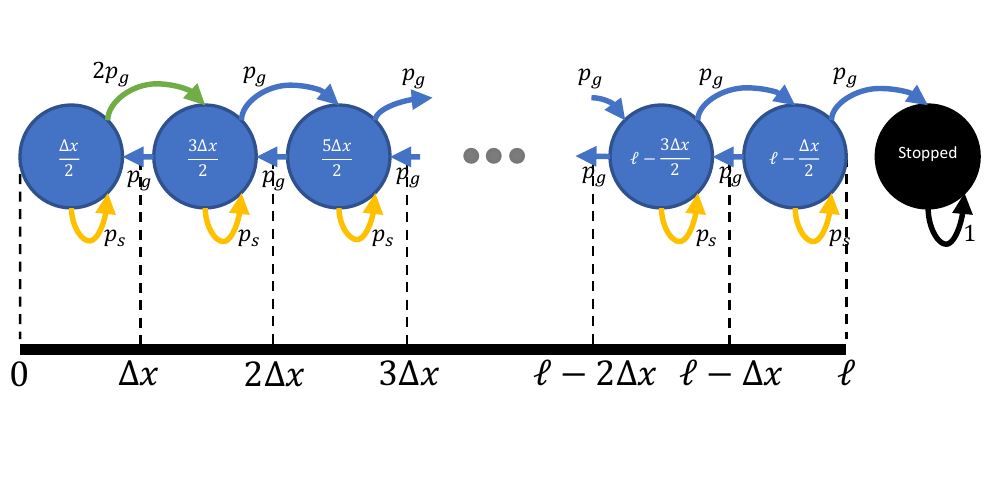}
\caption{Illustration of Markov random walk process obeying the law of $X_t$.  For all positions in the interior of the wire, a walker may move in a single timestep $\Delta t$ to the left or right with equal probability or stay in place.  Zero is a reflecting boundary and a walker ends its journey by stepping into the absorbing state beyond the length of the wire.}
\label{fig:monte}
\end{figure}

With these probabilities calculated, we can construct a Markov process as in Figure \ref{fig:monte}.  A random walker can move to adjacent points on the mesh with probability $p_g$ and stays in place with probability $p_s$.  If at zero, the reflecting nature of the walk forces it to the right with probability $2p_g$.  When the walker leaves the wire, it exits forever into an absorbing state.

To approximate the solution, we can use the following algorithm.
\begin{enumerate}
\item For a position $x_i$ on the mesh, initialize $M$ random walkers starting at position $x_i$.
\item Simulate each of the $M$ walkers, keeping track of the cumulative number $n_{ij}$ of walkers on node $x_j$ which began on node $x_i$.  End when all walkers have been absorbed.  \textbf{Do not include the initialization as part of the cumulative count.}
\item Repeat or parallelize for all positions $x_i$
\item Assign
\begin{equation}
\mathbb{E}\left[-F\int_0^T\ell-X(s)\,\rmd s\,\middle|\,X_0 = x_i\right]\approx-\frac{F\Delta t}{M}\sum_jn_{ij}\left(\ell - x_j\right):=u_i.
\label{eq:neuro_approx_1}
\end{equation}
\item Then,
\begin{equation}
u\left(x_i\right)\approx u_i - u_0.
\label{eq:neuro_approx_2}
\end{equation}
\end{enumerate}
Note, that the solution is being approximated at the midpoints of the divisions of the interval $[0,\ell]$, and not on the division points.  Hence, in the preceding equation, $u_0$ is calculated at the first midpoint from the left.  An estimate for how long to run each until all walkers are absorbed can be obtained through \eqref{eq:expected_stopping_time}.  This will only give the mean running time; a considerable percentage of the walkers will run longer.

\section{Simulation Results}
\label{sim_results}

\begin{figure}[h]
\centering
\includegraphics[width=3in]{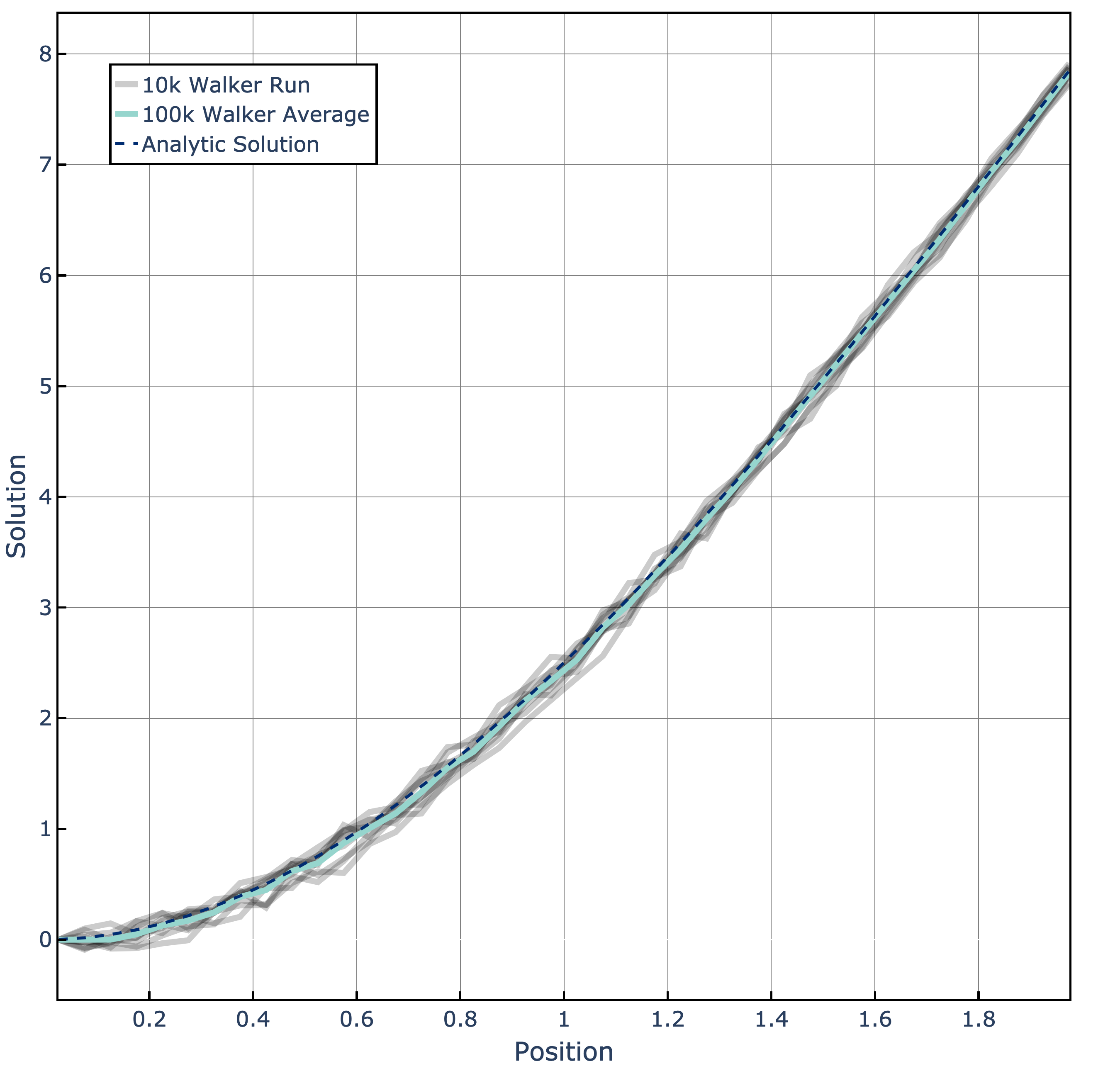}
\caption{Random walk approximation of \eqref{eq:heat_prob_sol} utilizing 10,000 walkers for each mesh point.  Gray lines showcase ten separate simulations and the teal line is the average of all ten simulations.  For comparison, the analytical solution \eqref{eq:heat_ana_sol} is plotted as a dashed blue line.}
\label{fig:comp_sim}
\end{figure}

To demonstrate this method, we simulated the previous algorithm using the values $F=3$, $\ell=2$, $\Delta x=0.05$, and $\Delta t= 0.0001$.  For this choice of $\Delta x$ and $\Delta t$, the probability of transitioning more than a single node is much less than $0.05$.  We simulated using $M=10,000$ random walkers for each mesh point.  The approximate solution obtained for ten separate simulations is shown in gray in Figure \ref{fig:comp_sim}.  The average of these ten runs is shown in teal.  The average can be interpreted in two ways.  We can either view the average (teal) as the empirical expected result of a simulation using $M=10,000$ walkers per mesh point with a visual range of uncertainty (gray); or, since all ten runs were distinct, the average (teal) could also be interpreted as a single run with $M=100,000$ walkers for each mesh point.

We then proceeded to create a spiking neural network implementation adapted from the density-based random walk method presented in~\cite{severa2018spiking}.  In this network, each node of the walk is represented by a sub-circuit of neurons that collects and distributes walkers according to the predefined probabilities (i.e.~$p_g$).  The spiking network performs several steps for each walker as the walker is counted and routed (though this is done in parallel for each node).  This time cost creates two separate time scales, and to avoid confusion, we will define two terms.  The phrase \textit{neural timestep} refers to one timestep of the spiking neural network algorithms, for example one `tick' on TrueNorth.  Explicitly, one neural time step is a cycle that involves every neuron's integration, fire, leak, and spike transport dynamics.  The phrase \textit{simulation timestep} refers to one timestep of the random walk process.  

Note that the relation between neural timesteps and simulation timesteps is not fixed as it depends on the number of walkers at each node.  In particular, the number of walkers on the node with the most walkers determines the number of neural timesteps required to evaluate a simulation timestep.  We recognize that one benefit of the neuromorphic approach is that we can create $n$ parallel tiles or copies of the network with $M/n$ walkers in each tile.  This reduces the number of neural timesteps required considerably and makes good use of large-scale neural systems.  

Finally, in our ideal random walk simulation, we run all the walkers until they reach the absorption node.  However, with current interfaces to neuromorphic hardware, this is often difficult, and instead we use a large, predetermined number of neural timesteps.  Any walkers that have not been absorbed by the end of the simulation increase the error.  It may be possible to estimate the number of neural timesteps required a priori though we have not completed this analysis.  

\begin{figure}[h]
\centering
\includegraphics[width=3in]{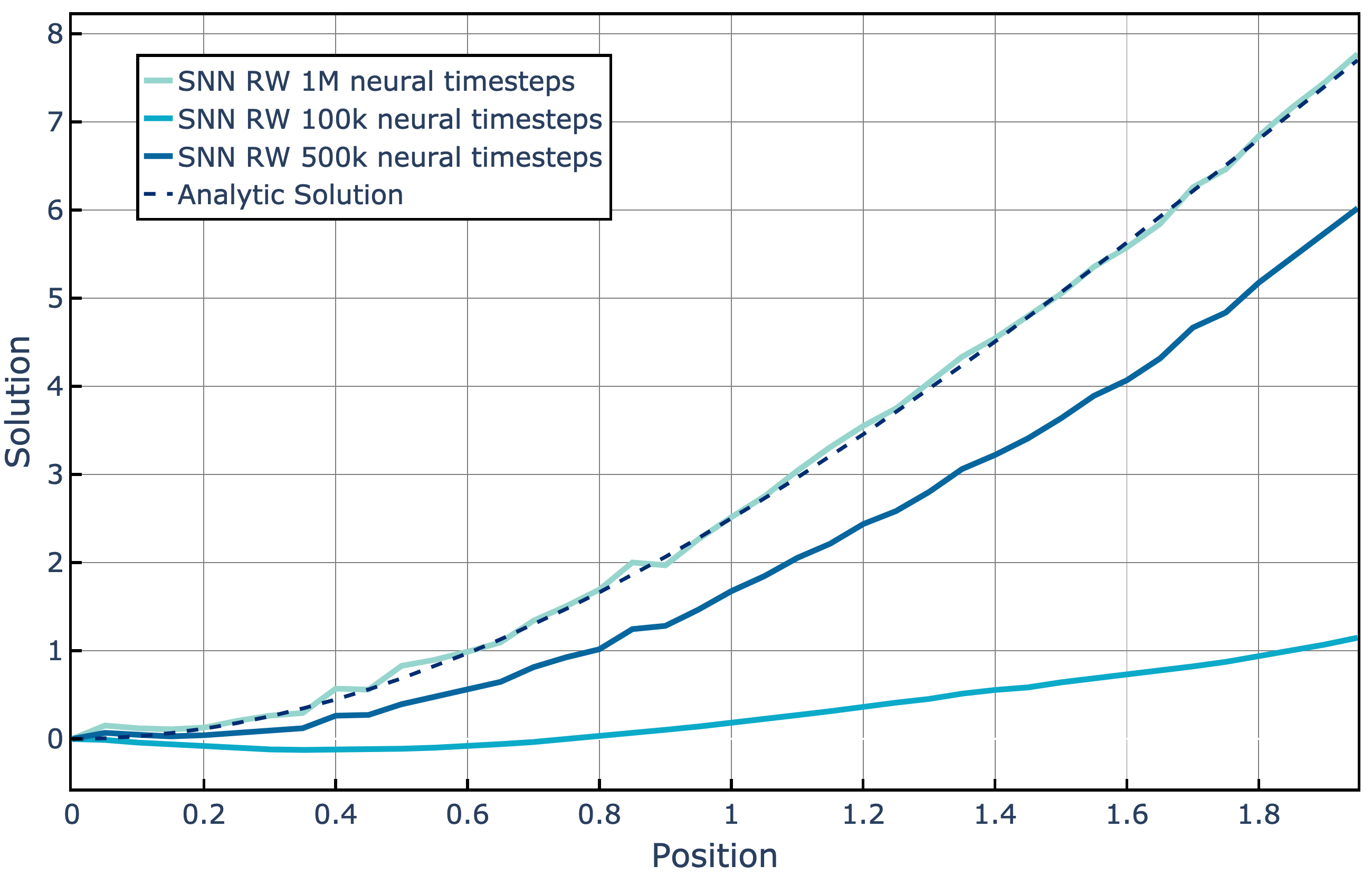}
\caption{ Random walk approximations using a spiking neural network simulator at various lengths of simulation.  Each line represents the results of 10,000 walkers.   \label{ds_results}}
\end{figure}

Figure~\ref{ds_results} shows the results of a software simulation of the spiking neural network.  As expected, increasing the number of neural timesteps (thereby increasing the number of simulation timesteps) improves the approximation considerably.  The listed results are from 100 tiles of 100 walkers which is equivalent to a simulation with 10k walkers.  To increase performance, walkers that reach the absorption node are removed from the simulation.  

\begin{figure}[h]
\centering
\includegraphics[width=3in]{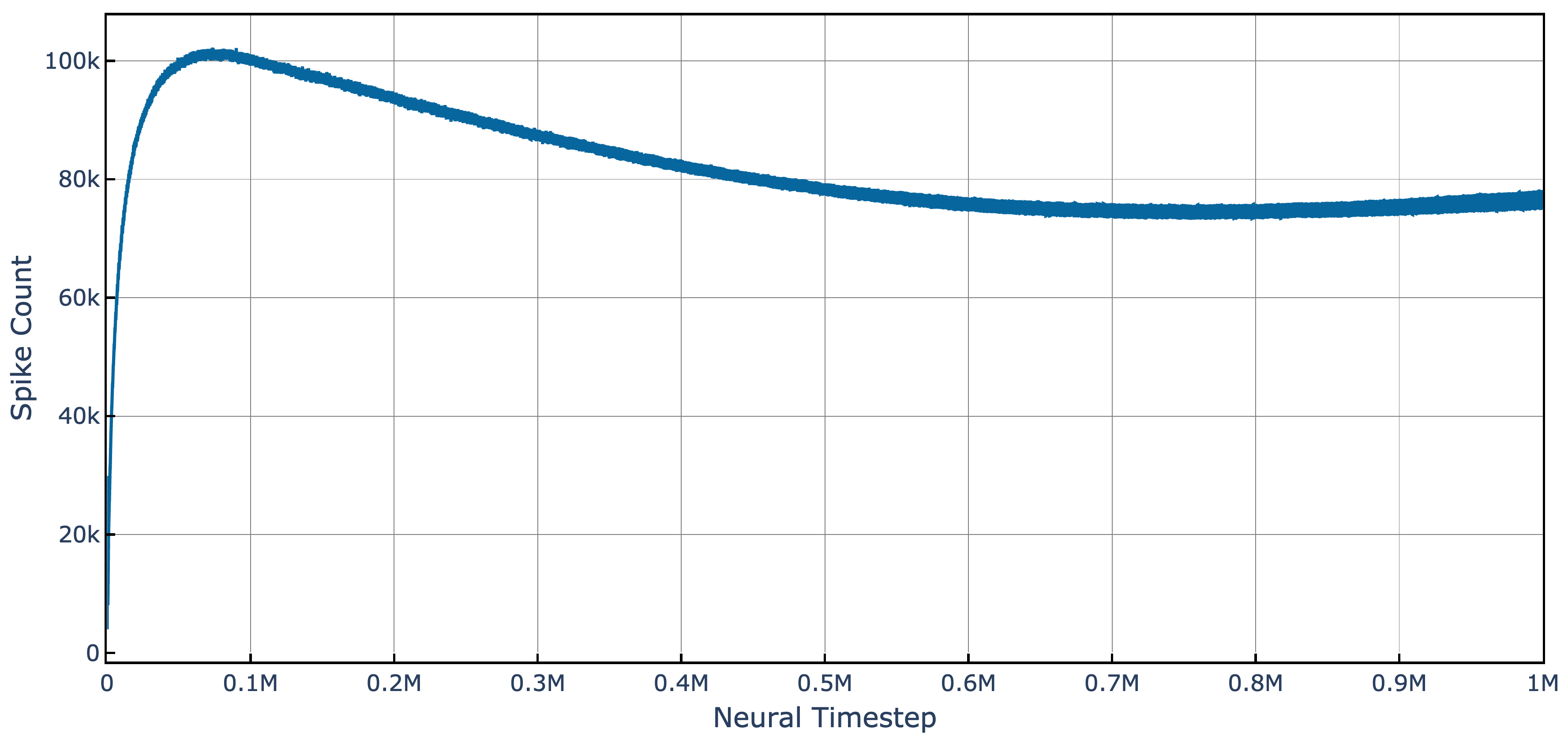}
\caption{The moving average (25 timestep window) of spikes in flight at each timestep during a simulation of 1M neural timesteps with 10k walkers distributed evenly over 100 tiles. Results are aggregated across all the starting node positions and all tiles. Each tile requires approximately 400 neurons; the total number of neurons is approximately 1.6M.\label{spike_figure}}
\end{figure}

To help quantify the load on a neuromorphic system, we analyzed the number of spikes in flight at each timestep, see Figure~\ref{spike_figure}.  Like the results in Figure~\ref{ds_results}, the simulation was divided into 100 tiles of 100-walker runs for an effective 10k walker run.  Interestingly, the moving average shows that the load increases suddenly in the early part of the simulation but quickly reaches a maximum.  After the maximum, the spike counts slightly decrease.

\section{Hardware Results}
\label{hardware_results}

We implemented this algorithm on both the IBM TrueNorth~\cite{merolla2014million} and Intel Loihi~\cite{davies2018loihi} Neuromorphic systems.  The spiking network is relatively simple, with relatively low connectivity requirements; however, the stochastic component benefits from a platform dependent formulation.  Functionally, this collection of neurons must take a walker (represented by a spike) and route it to a number of output neurons according to a mutually exclusive probability draw.  There are several ways to accomplish this functionality depending on the neuron model used and the way each platform implements stochasticity.  Because Loihi and TrueNorth have different stochastic neuron models, our implementations of the circuits between the two platforms are different, but functionally equivalent.  Since the simple structure of the underlying example problem requires only 3 neighboring nodes, the difference in the time and resource costs between the two stochastic implementations is low. Additionally, both Loihi and TrueNorth have limited precision in their random number generators, which required some care in configuring these circuits, and likely explains some of the differences between our neuromorphic results and the analytical solution, which assumes a much greater precision in the probabilities.

\subsection{Loihi Results\label{Loihi_results}}

We deployed our neural circuit onto an 8 Loihi-chip Nahuku neuromorphic platform. As this model is relatively compact in size, we fit a single instance onto three cores: one for model supervision, one for deterministic mesh points, and one for the stochastic neurons. All simulations were performed in serial to enable benchmarking, however in principle the 8 Loihi chips could run several hundred copies of this network in parallel.
		
\begin{figure}[h]
\centering
\includegraphics[width=3in]{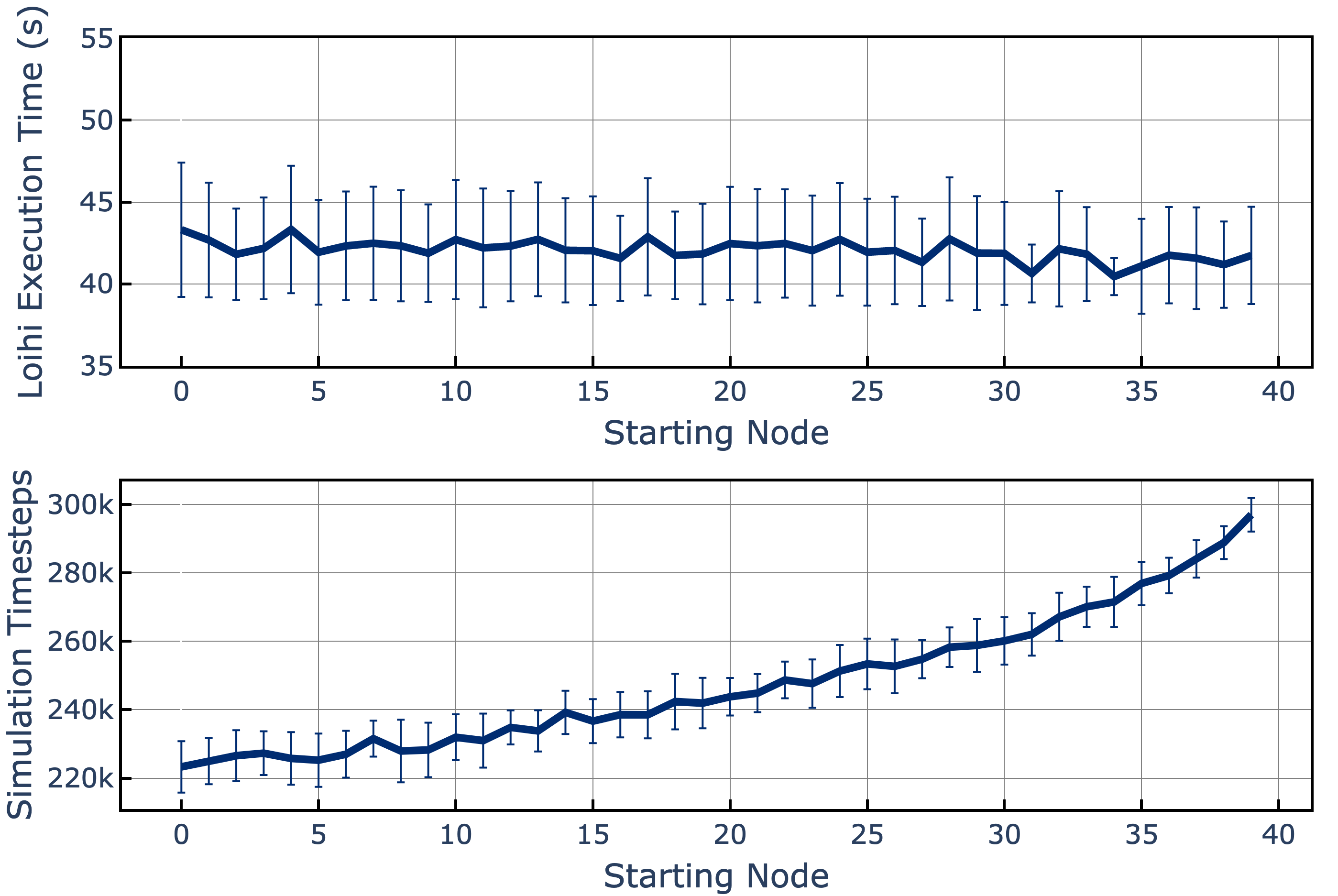}
\caption{Simulation of wire heat distribution on Loihi. Execution time was relatively constant for 7.5 million neural updates with 250 walkers starting at different locations (top).  While execution time was constant, the number of simulation timesteps varied considerably from left to right for the fixed number of neural timesteps due to walkers being removed from the simulation at the right side of the wire (bottom).  For both, plotted errors are standard deviations ($\mathbf{N=40}$).\label{loihi_time}}	
\end{figure}

Simulation execution time on Loihi was constant across starting locations, taking 42 seconds on average to simulate 250 walkers over 7.5 million neural timesteps (Figure~\ref{loihi_time}, top). The number of simulation timesteps (i.e., number of wire model updates) did vary considerably according to starting location (Figure~\ref{loihi_time}, botom).  This is because simulations whose walkers start on the right side near the sink were removed from the simulation faster, thus speeding up overall execution time. Overall execution time and number of simulation timesteps scaled linearly with number of neural timesteps (data not shown).  

\begin{figure}[h]
\centering
\includegraphics[clip=10 10 10 10,width=3in]{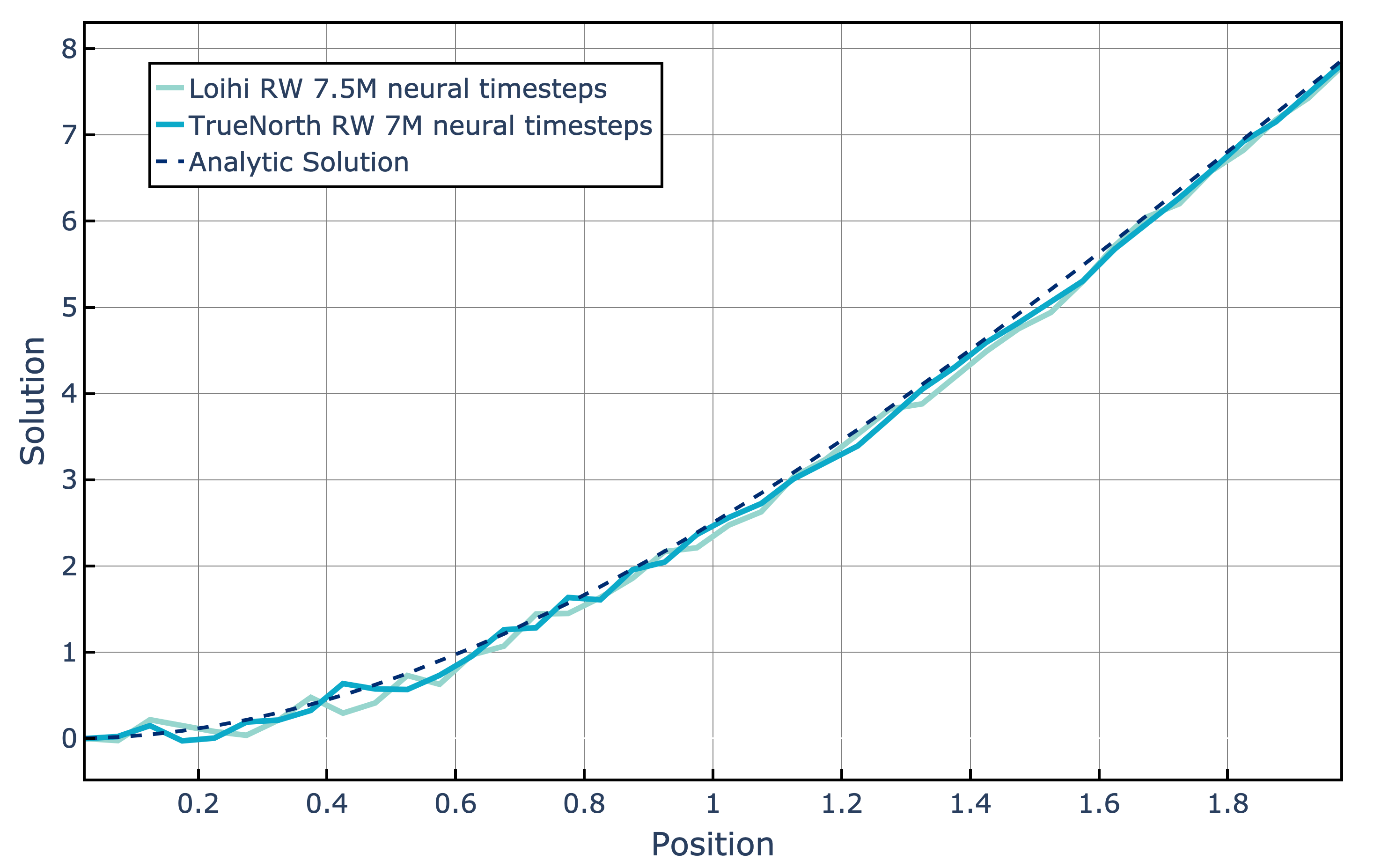}	
\caption{\label{hw_plot}Random walk approximations for 10k walker study on TrueNorth and Loihi.}
\end{figure}

Figure~\ref{hw_plot} shows results from Loihi considering 10,000 walkers starting on each wire location. These simulations were performed in batches of 250 walkers each, though this number could be increased with minor changes to the network. Overall, the Loihi results match the analytical solution reasonably well, remaining within the variation captured in Figure \ref{fig:comp_sim}.
 
As has been recognized with neuromorphic platforms, I/O is potentially a bottleneck for applications. As these are contained numerical simulations, I/O can be limited to the setup of the simulation and offloading of cumulative results. However, Loihi does limit the number of probes that can be used to monitor activity, which would have to be considered in models with more grid points. Further, continuously reading out neural activity during the simulation slows the simulation dramatically, so in order to minimize I/O, the cumulative counts of walkers were stored in the voltage of a set of no-decay neurons that were only read-out at the end of the simulation.

\subsection{TrueNorth Results}
\label{TrueNorth_results}

We implemented the density method of the spiking random walk algorithm onto a single chip of the TrueNorth hardware. Taking advantage of parallelism, we make 1,000 tiles of the simulation with each tile having 10 walkers. This results in a combined simulation of 10,000 walkers. This implementation required 4,067 of the 4,096 cores of the TrueNorth hardware. We ran the hardware for 7 million neural timesteps. Figure~\ref{hw_plot} shows the solution results of this simulation starting at each wire location, which match the analytical solution very closely.

\begin{figure}[h]
\includegraphics[width=3in]{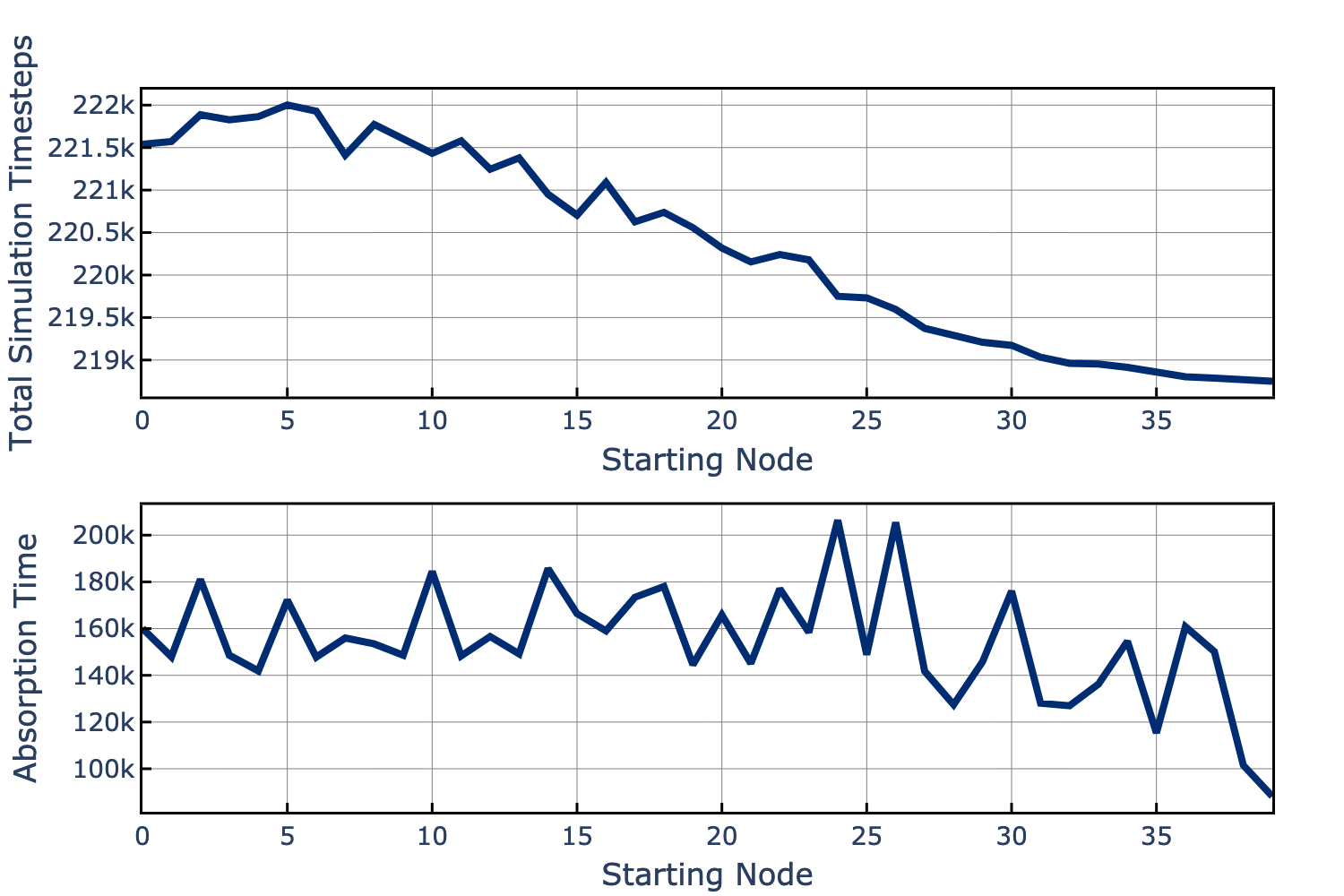}
\caption{\label{tn_timing} Timing results for simulations on TrueNorth: The total number of simulation timesteps for a 7 million neural timestep study decreases as the starting node approaches $\mathbf{\ell}$ (top).  The number of simulation timesteps required for all walkers to leave the system varies considerably, though consistently across starting nodes before ultimately dropping near $\mathbf{\ell}$ (bottom).}
\end{figure}

Figure~\ref{tn_timing} illustrates data on the number of simulation timesteps produced from the $7$ million neural timesteps and the number of simulation timesteps each starting location took to reach full absorption. Figure~\ref{tn_timing} (top) produces a different trend than observed on Loihi (Figure~\ref{loihi_time}, bottom) due to a difference in implementation. On TrueNorth we did not remove any walkers once they reached the sink. Thus, as walkers accumulate in the sink it requires more neural timesteps per simulation timestep. Starting positions closer to the sink accumulate walkers earlier in the simulation, increasing the ratio of neural to simulation timesteps, and driving down the total number of simulation timesteps produced from the fixed 7 million neural timesteps. The average number of neural timesteps per simulation timestep was 31.8 with a standard deviation of 0.166. The absorption time (Figure~\ref{tn_timing}, bottom) varied greatly per starting location but showed a downward trend towards the sink node. This is expected since starting locations next to the sink would on average absorb more walkers early in the simulation.

Timing analysis is not readily available on the TrueNorth platform. The function call that executes the model in hardware includes model load time, execution time, and spike I/O. Therefore, to derive exact execution time we run three different iterations of the model. The first iteration executes the model for a single hardware cycle. This tells us how long it takes to load the model since no output spikes are produced after just one neural timestep. The second iteration runs the full 7 million neural timesteps but has spike I/O disabled. This tells us the time for loading the model plus executing the model. Since the first iteration provides us model load time, we can subtract that out of the second iteration's time and arrive at model execution time. The third iteration runs the model for 7 million neural timesteps with spike I/O enabled. Removing the measured time of the second iteration from that of the third produces the resultant spike I/O time. Our reported timing is taken from the average over 10 executions of each iteration. 

We ran this timing analysis on a simulation of a single starting position in the middle of the wire. Additionally, we configured the tick rate of the TrueNorth hardware to be as fast as possible for this model. The execution time was 3,386 seconds for 7 million neural timesteps, which equates to a tick rate of approximately 484~$\mu s$ per tick, or approximately 2~kHz. The simulation produced 11.25 billion spikes that were ex-filled in 834 seconds, a spike rate of approximately 13.5 million spikes per second.

In addition to the I/O bottlenecks discussed previously, hardware may impose additional difficulties, limiting the ability to assign probabilities accurately.  For TrueNorth, the hardware expression of the stochastic parameter of a neuron has a resolution of $1/256$. On a problem-specific basis, this could cause wild and significant error with slight changes.  Focusing discussion on this problem, recall \eqref{eq:neuro_approx_1} and \eqref{eq:neuro_approx_2}.  For any $x_j\in[0,\ell]$, the interior of the sum in \eqref{eq:neuro_approx_1} is positive, forcing the entire value of $u_i$ to be negative for any $i$.  For a fixed $i$, the value of $u_i$ is decreased most strongly if the underlying random walks congregate more toward the left end of the rod.  That is, if the values of $n_{ij}$ are higher for those indices corresponding to locations closer to zero.

\begin{figure}[h]
\centering
\includegraphics[clip=10 10 10 10,width=3in]{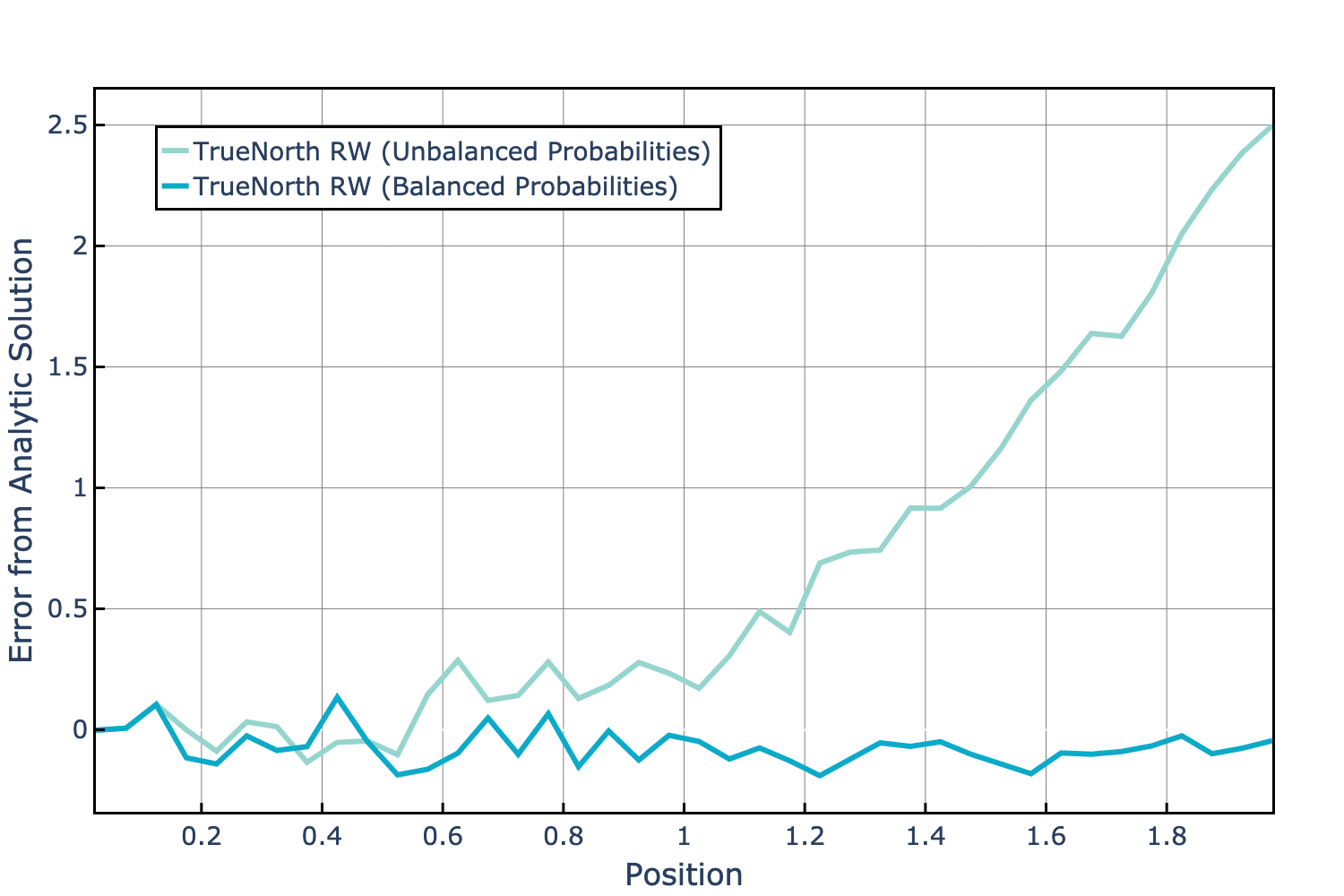}	
\caption{Error resulting from an imbalance in probabilistic values expressed in hardware\label{tn_error}}
\end{figure}

Initially, due to a one-sided round in our corelet code, our TrueNorth network approximated $p_s \approx 0.9258$ while $p_g$ was different for left and right transition.  The effective probabilities were $p_\text{left}\approx0.0374$ and $p_\text{right}\approx 0.0368$.  The true values for the parameters of simulation should be $p_s \approx 0.9229$ and $p_\text{left}=p_\text{right}=p_g\approx0.0385$.  While the TrueNorth values may not appear too different from the exact values, a problem arises because $p_\text{left}>p_\text{right}$.  Intuitively, this imbalance means a random walk is more likely to move towards the left than the right.  This will cause higher values of $n_{ij}$ than normal for locations closer to zero.  This will increase the negative values of $u_i$ causing the estimated temperature \eqref{eq:neuro_approx_2} to be higher than normal, see Figure~\ref{tn_error}.  An underestimation of the true temperature would occur if the probability of transition favored the right.  Therefore, for this problem it is extremely important to ensure that the probability of transition to the left and the right is equal.

\section{Evaluating Neuromorphic Hardware}
\label{eval_intro}
Developing a methodology for benchmarking neuromorphic systems has become critical for the success of the field.  Several efforts have been undertaken to either benchmark specific systems or call out larger perspectives on neuromorphic benchmarking~\cite{diamond2016comparing, liu2016benchmarking, davies2019benchmarks}.  Regardless, this is still a critical open question that faces the field.  Benchmarking across platforms often requires concessions due to network incompatibilities. Without well-adopted specifications, it is difficult to obtain the appropriate data and software as well as run the benchmark itself.  While any single task will not sufficiently determine the performance of a system, we suggest that this random walk/steady-state heat equation (Section \ref{steady_state_section}) offers a compelling benchmark task.  When viewed as a benchmark task, this algorithm has several desirable qualities:
\begin{enumerate}
\item The task does not rely on outside data, and so it is fully self-contained.
\item The steady-state heat equation has an easy-to-grasp analytic solution, and so results of the benchmark are directly comparable.
\item Requirements on the neuron model are simple, allowing this to target many hardware platforms.
\item The algorithm scales both in the number of nodes (neurons) and the number of walkers (spikes), which provides a interaction between space and activity costs.
\item The majority of connectivity is local, and the activity is relatively sparse.
\item The algorithm requires a simple pattern be repeated across the fabric.
\end{enumerate}

Given that the analytical solution is easily computable, the solution approximated by the simulation is not the interesting metric.  Instead, this task can be used to evaluate the performance of a system in terms of computational speed and scalability.  As shown in Section \ref{sim_results}, the rate of spikes is relatively low, but varies over time, which can help characterize a system's spike routing ability.  Additionally, if possible, practitioners can compare running energy costs across different neuromorphic platforms.  Rather than a single value, for both performance and efficiency, we suggest reporting results relative to the simulation timestep and starting node (details in Section \ref{neuro_implementation}) as loads depend on both of these factors.  Scaling various parts of the algorithm can be used to test specific hardware capabilities:
\begin{enumerate}
\item The number of tiles tests the parallelism.
\item The number of walkers tests the spike throughput.
\item The number of neural timesteps tests performance and I/O limits.
\end{enumerate}
However, we recognize that it is difficult to report a true apples-to-apples across systems: software interfaces for different platforms have different capabilities;  power consumption estimates may be dependent on research agreements; I/O may carry a prohibitively high cost.  Because of these challenges, reporting any single metric becomes disingenuous, and we suggest that if used as a benchmark task, a full picture of system performance is reported.

To support this benchmarking effort, we have developed a short Python script that will generate a network for solving this equation, parameterized by the number of walkers and the number of nodes in the system. The script produces either a NetworkX DirectedGraph object or file with the specified neuron properties and synapses. This script is posted online at (url determined after the acceptance of the paper).

\section{Conclusion}

The task described here is an example of how neuromorphic hardware can have an impact on a much broader set of numerical applications than the community generally considers. Demonstrating the ability for spiking neuromorphic systems to impact conventional numerical computing is important; by extending its impact beyond cognitive applications we increase the likelihood of a long-lasting effect on the computing field.  Notably, while we did not exert considerable effort on optimizing the results presented here for either time or space, we already have evidence that neuromorphic hardware can be more efficient than conventional approaches when fully parallelizable Monte Carlo based are implemented.  For example, with only minimal additional work and ignoring I/O considerations, we anticipate that we could potentially perform the complete simulations described above in parallel on a fully populated 32-Loihi chip Nahuku board in less than a minute. 

Not surprisingly, we observed several aspects of neuromorphic hardware that will require further investigation. For one, the I/O costs of neuromorphic hardware will likely grow as a consideration.  The costs of I/O are often considered for streaming applications such as real-time machine learning inference, but for the class of numerical simulations considered here, interactive I/O is not required, but tracking state---or in this case accumulating state---is required in order to properly evaluate the simulation. Since I/O will likely continue to be a limiting factor, further processing of simulation outputs on the neuromorphic substrate is likely ideal.  One way to do this is to implement the post-processing steps that we performed offline as neural circuits themselves and integrate them into a fully composed simulation~\cite{aimone2019composing}.

The second significant consideration learned from the neuromorphic simulations is the potential impact of reduced precision stochastic neurons on model performance. The stochastic steps of our simulation are affected by the precision of internal neuron states, precision of weights between neurons, and precision of the random number generator. These different components interact in complex, architecture-dependent ways and the implications of this reduced precision merit deeper exploration. At the same time, some of the benefits of neural hardware---the ability to have more random number draws effectively in parallel---may be able to offset these consideration.

Nevertheless, these neuromorphic considerations should prove surmountable especially as future generation platforms become available. As non-anticipated applications such as these are explored, it will be increasingly evident what the potential implications of reduced precision and I/O are and whether the costs of mitigation advocate for future hardware modifications or improved circuit and algorithm design.

\section{Acknowledgments}
This work was supported by Laboratory Directed Research and Development program at Sandia National Laboratories. Sandia National Laboratories is a multi-program laboratory managed and operated by National Technology and Engineering Solutions of Sandia, LLC., a wholly owned subsidiary of Honeywell International, Inc., for the U.S. Department of Energy's National Nuclear Security Administration under contract DE-NA-0003525. This paper describes objective technical results and analysis. Any subjective views or opinions that might be expressed in the paper do not necessarily represent the views of the U.S. Department of Energy or the United States Government.  SAND2020-5296 O

\bibliographystyle{agsm}

\end{document}